# Robust Cross-vendor Mammographic Texture Models Using Augmentation-based Domain Adaptation for Long-term Breast Cancer Risk

Andreas D. Lauritzen[a,*], My Catarina von Euler-Chelpin[b], Elsebeth Lynge[b], Ilse Vejborg[c], Mads Nielsen[a], Nico Karssemeijer[d], and Martin Lillholm[a]

[a]University of Copenhagen, Faculty of Science, Department of Computer Science, Copenhagen, Denmark
[b]University of Copenhagen, Faculty of Health and Medical Sciences, Department of Public Health, Copenhagen, Denmark
[c]Gentofte Hospital, Department of Breast Examinations, Gentofte, Denmark
[d]ScreenPoint Medical, Nijmegen, the Netherlands

**Abstract**

**Purpose:** Risk-stratified breast cancer screening might improve early detection and efficiency without comprising quality. However, modern mammography-based risk models do not ensure adaptation across vendor-domains and rely on cancer precursors, associated with short-term risk, which might limit long-term risk assessment. We report a cross-vendor mammographic texture model for long-term risk.

**Approach:** The texture model was robustly trained using two systematically designed case-control datasets. Textural features, indicative of future breast cancer, were learned by excluding samples with diagnosed/potential malignancies from training. An augmentation-based domain adaption technique, based on flavorization of mammographic views, ensured generalization across vendor-domains. The model was validated in 66,607 consecutively screened Danish women with flavorized Siemens views and 25,706 Dutch women with Hologic-processed views. Performances were evaluated for interval cancers (IC) within two years from screening and long-term cancers (LTC) from two years after screening. The texture model was combined with established risk factors to flag 10% of women with the highest risk.

**Results:** In Danish women, the texture model achieved an area under the receiver operating characteristic (AUC) of 0.71 and 0.65 for ICs and LTCs, respectively. In Dutch women with Hologic-processed views, the AUCs were not different from AUCs in Danish women with flavorized views. The AUC for texture combined with established risk factors increased to 0.68 for LTCs. The 10% of women flagged as high-risk accounted for 25.5% of ICs and 24.8% of LTCs.

**Conclusions:** The texture model robustly estimated long-term breast cancer risk while adapting to an unseen processed vendor-domain and identified a clinically relevant high-risk subgroup.

## 1    Introduction

Breast cancer is the leading cause of death in women worldwide.[1] Approximately 13% of woman in the E.U. and U.S. will receive a breast cancer diagnosis in their lifetime.[2,3] Population-based screening by mammography, an x-ray of the compressed breast, succeeds in decreasing mortality by detecting early breast cancer signs.[4] Screening programs often employ a one-size-fits-all approach where all women are screened under equal conditions. This approach is inefficient considering the low breast cancer incidence and general lack of specialized radiologists. Personalized screening has the potential to improve screening by reallocating radiologists' resources efficiently to high-risk women without increasing the total cost or compromising screening quality.[5,6]

M. Gail et al. showed that breast cancer risk could be estimated using established clinical risk factors such as family history of breast cancer, reproductive history, and history of previous biopsies.[7] The Tyrer-Cuzick (TC) model, extended the Gail model to incorporate genetic information, e.g., BRCA mutations in first- and second-degree relatives.[8] Whereas models based on clinical risk factors can estimate risk reasonably well, the extensive data collection and potentially inconsistent self-reporting makes implementation in population-based screening programs laborious and complex.

Conversely, mammography-based risk models derive information directly from the already available screening mammogram. Mammography-based classification schemes like the Wolfe and Boyd models showed that the amount of radio-dense tissue, e.g., lobules and ducts, relative to the amount of fatty tissue could successfully categorize high-risk women.[9,10] This ratio between dense and fatty tissue, the percentage mammographic density (PMD), became an established risk factor. L. Tabár et al. later showed that the distribution of dense/parenchymal breast tissue, not solely the amount, was a risk factor as well.[11] The study showed that mammographic features, representing breast heterogeneity, could be manually scored to identify high-risk women. Such features are often referred to in literature as mammographic texture. Mammographic density and texture are global features of the mammogram associated with long-term breast cancer risk. This is contrary to detection of actual cancerous lesions, minimal signs of developing cancer (precursors), or other localized findings which are associated with short-term breast cancer risk. Furthermore, women recalled at screening with suspicious findings, but with a negative result for breast cancer, have an elevated long-term risk as well.[12]

Modern mammography-based risk models automatically derive breast cancer predictive features such as precursors, texture, or density, and rely on supervised deep learning (DL) methods trained on large datasets of labeled digital mammograms. DL risk models are more compatible with population-based screening programs as digital mammograms can be made available for analysis immediately after or during screening. DL risk models relying on precursors or breast density are currently being evaluated in real and simulated clinical settings.[5,6] However, long-term risk models have not been evaluated in terms of clinical utility.

We distinguish between three approaches to estimate mammography-based breast cancer risk by DL methods. Firstly, a diagnostic model, trained to detect or segment suspicious local findings, can be used to estimate short-term risk.[13,14] Secondly, a texture model can be trained to estimate long-term risk by learning systemic differences in tissue composition between high- and low-risk women, e.g., mammographic texture or density patterns indicative of elevated risk.[15,16] Thirdly, a conflated model can be trained to simultaneously estimate immediate, short-, and long-term risk by learning both features of visible cancers/precursors, and systemic/global differences in mammographic texture.[17,18] It has, however, been suggested that a visible lesion is an easily distinguishable feature but is weakly predictive of long-term risk and therefore might prevent an DL model from learning strongly predictive features of long-term risk.[16,19]

Some of the previous studies were performed with mammograms in raw format, i.e., the untouched output from a mammographic device directly representing x-ray attenuation.[6,15] Raw data is well-defined and mostly homogeneous across vendors and mammographic devices, which makes it easier to develop robust AI algorithms. However, models that rely on availability of raw data have limited clinical value, because clinics use processed mammograms prepared by the mammography device with a contrast suitable for human visual interpretation. Raw mammograms are normally not archived whereas processed mammograms are routinely archived. The appearance of processed views depends highly on the vendors image post-processing pipelines and even depends on the screening clinics' own preferences/settings. Consequently, DL risk models must not only estimate risk in processed views but also robustly across mammographic devices from different vendors to remain clinically relevant at a large scale. Most of the relevant work we compared to considered only processed mammograms from a single vendor.

Modern DL models often suffer from domain shift problems and will in many cases of medical images not adapt to unseen domains, e.g., to processed mammograms obtained on a mammographic device from another vendor than the one used for training. The ability to adapt to unseen vendor-domains would allow retrospective external validation at multiple international clinics and minimize clinical implementation overhead. Device-domain adaptation for DL breast cancer risk models have not been investigated in depth yet.

In this study, we developed a clinically relevant, robust, cross-vendor texture model for long-term breast cancer risk. The main contributions of this study were as follows:

1. We trained an DL model, based on a well-validated architecture, for texture long-term breast cancer risk using screen-available information only. To ensure optimal training and not depressing a potentially weak long-term risk signal, we train exclusively on views with no visible or potentially visible diagnosed cancers.

2. We showed that training for long-term risk can produce erratic convergence which we alleviated by excluding noise-inducing training samples and systematically designing a case-control training dataset for an ensemble texture model.

3. To ensure adaptation across vendor-domains, we trained the texture model using a data augmentation technique converting raw to processed views in multiple flavors. The texture model was successfully applied to domains of flavor-processed and vendor-processed views.

4. The texture model was shown to generalize well across populations and vendors in two large independent screening datasets and identified clinically relevant high-risk women.

## 2   Related Work

Breast cancer risk is used in different contexts throughout the literature and is often intertwined or confused with detection/segmentation. It is important to consistently distinguish between models created purely for detection, risk assessment, or a combination and report results accordingly. In the following sections, we describe existing methods for estimating breast cancer risk: a diagnostic-based, a texture-based, and two conflated models.

### 2.1  Diagnostic Models for Risk Estimation

A. Gastounioti et al. employed a diagnostic model, the ProFound AI tool (iCAD), which detects/segments suspicious lesions and was compatible with Hologic, GE, and Siemens devices. The model included age, masses, microcalcifications, asymmetry, and breast density to provide a short-term risk estimate in a cohort of screen negative women with two-year follow-up. The model achieved an area under the receiver operating characteristic curve (AUC) of 0.68.[13] No long-term risk estimates were presented. There exist other studies using a similar approach to estimate short-term risk.[14] We focused on a long-term risk model by excluding localized findings from training as suggested by Liu et al..[16]

*2.2 Mammographic Texture Model for Risk Estimation*

Kallenberg et al. trained an autoencoder on multi-scale image patches to learn mammographic features in an unsupervised fashion. The model was finetuned to measure texture risk, given only cancer contra-lateral views, assuming tissue patterns persisted across left and right breasts.[15] The model achieved an area under the receiver operating characteristic curve (AUC) of 0.61 in a dataset with two years of follow-up. The training and validation datasets were small, from the same population, and single-vendor raw views only.

*2.3 Conflated Models for Risk Estimation*

Dembrower et al. trained an Inception-ResNet-v2 model to estimate risk on multi-scale image patches of mammograms. The model included age and device acquisition parameters.[17] The model was trained and validated on women from one population with Hologic-processed views only who was cancer-free for 12 months and achieved an AUC of 0.65. The training views might include early or slow-growing malignancies thus yielding a conflated model. There was no evaluation of robustness regarding early stopping criteria.

Yala et al. trained a conflated model, named Mirai, to estimate one to five-year risk.[18] The model consisted of a ResNet encoder, a transformer model to aggregate features, a risk factor predictor, and finally an additive-hazard layer to estimate risk at multiple timepoints. Mirai was trained using an adversarial approach to learn device-invariant features, thus being able to generalize across mammographic devices. However, only two systems from the same vendor (Hologic) were considered. Mirai achieved an AUC between 0.68 and 0.73 in seven independent datasets, all with Hologic mammograms. However, during validation, 42 to 82% cancers were diagnosed with one year and 54 to 91% within two years which might not be adequate to measure the long-term risk estimation performance and favoring Mirai's conflated approach.[20]

None of the above-mentioned studies investigated ability to adapt to unseen vendors, e.g., from Siemens to Hologic. As presented above, the problem of learning features of long-term risk in a data-driven fashion is inherently difficult and noise-filled. When approached as a supervised learning task with per-study labels, a modern DL model it is very likely to overfit by learning long-term risk-irrelevant features due to weak long-term risk signal compared to other visible features and model overparameterization.

Additional noise is present as high-risk women, e.g., high density and texture-risk women, might not develop a breast cancer within follow-up or at all. This can lead to erratic convergence and have not been discussed in literature in this context. Furthermore, existing methods are conflated models in terms training, but also mix detection of future/current cancers at different intervals during validation. We aimed to accommodate these above-mentioned limitations.

## 3 Materials and Methods

In the following sections, we describe our methods for developing and testing the texture model. Initially, we define four datasets – two for training and two for testing. Next, we describe our efforts to stabilize training. Then, we describe the augmentation-based domain adaptation technique to enable a cross-vendor texture model. Finally, we describe the training procedure and the ensemble texture model.

This study was approved by The Danish Data Inspection Agency and Danish Patient Safety Authority including collection and analysis of relevant data. The need for informed consent was waived.

*3.1 Screening Populations and Datasets*

The Danish screening population was from the breast cancer screening program in the Capital Region of Denmark, where women between the age of 50 and 70 were screened biennially on Mammomat Inspiration systems (Siemens Healthineers).

The Dutch screening population was from the breast cancer screening program in Utrecht, the Netherlands, where women between the age of 50 and 74 were screened biennially on Lorad Selenia systems (Hologic). Screened women had four mammographic views captured, two for each breast. The mammograms were read by two specialized radiologists who determined whether to recall the women for diagnostic tests. All breast cancers were diagnosed as invasive or ductal carcinoma in situ based on clinical examination consisting of mammography, ultrasound, and needle biopsy.

Breast cancers were separated into three distinct groups: Screen-detected cancers (SDC), interval cancers (IC), and long-term cancers (LTC). Recalled women who received a breast cancer diagnosis within six months from screening had SDCs. Women who did not have an SDC yet received a breast cancer diagnosis within 24 months from screening had ICs. LTCs are breast cancers diagnosed at least two years from screening regardless of recall. In practice, LTCs were SDCs or ICs in subsequent screening rounds.

From the two screening populations, two training datasets and two testing datasets were collected. Fig. 1a and 1b depict the collection processes in a flow diagram including the inclusion/exclusion process for women with visible artifacts, corrupted views, previous breast cancer diagnoses, and biopsy clips in the cancer contra-lateral breast or both breasts.

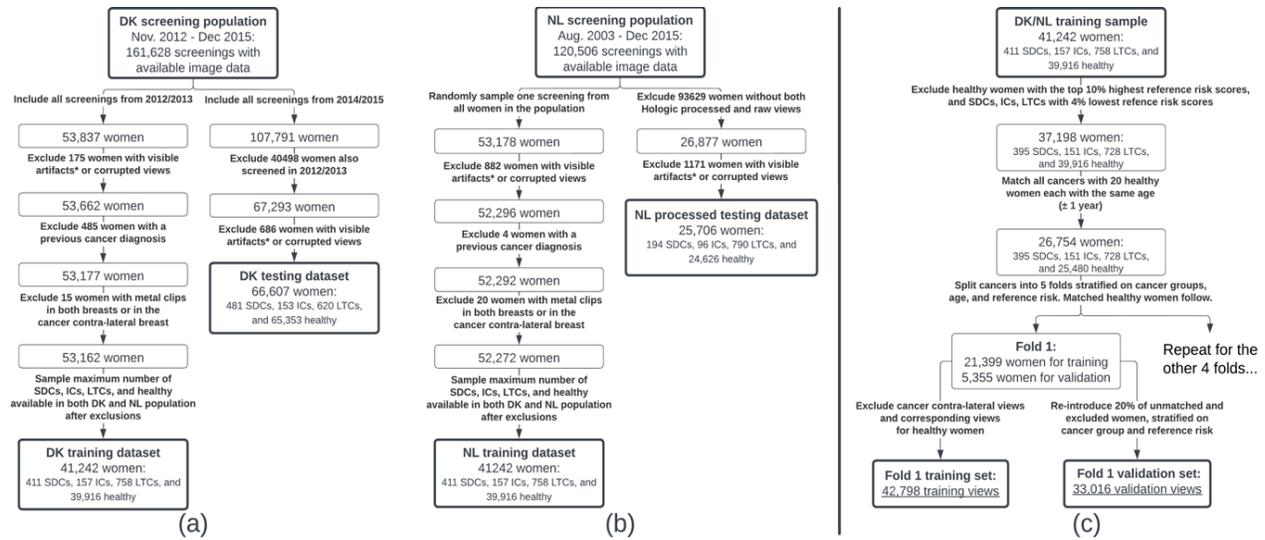

**Fig. 1 (a)** Flow diagram of selection process for the Danish training and testing dataset, **(b)** Selection process for the Dutch training and processed testing dataset, and **(c)** curation of views for training and validation in the filtering stage. *Visible artifacts include implanted medical devices, foreign unknown objects, and other disruptive image artifacts.

The Danish training dataset was collected from the Danish screening population as screened women from November 1st, 2012, to December 31st, 2013. The Dutch training dataset was collected from the Dutch screening population by randomly sampling a single screening for each woman between August 1$^{st}$, 2003, and January 1$^{st}$, 2015. After exclusions, the maximum amount of SDCs, ICs, LTCs, and healthy women were sampled from the Danish and Dutch training population, such that each group of women were equal in size. Both datasets consisted of raw mammographic views.

The Danish testing dataset was collected from the Danish screening population as screened women from January 1st, 2014, to December 31st, 2015. The Danish testing dataset consisted of raw mammographic views. This dataset was a hold-out testing dataset to measure performance for models trained on the Danish or Dutch training datasets.

The Dutch processed testing dataset was collected from the Dutch screening population between August 1$^{st}$, 2003, and January 1$^{st}$, 2015, and was defined as women for which raw and corresponding Hologic-processed views were available. The two Dutch datasets were not independent, but this was not a limitation because the Dutch processed testing dataset was only used for evaluation of texture models trained on Danish data.

*3.2 Curation of Training Data to Stabilize Convergence*

Due to noisy labels and the inherent difficulty of the task, the reference training of the texture model was erratic with unstable convergence (see Fig. 5). To attain reliable risk estimates, we curated a systematically designed case-control dataset for robust training. This curation process consisted of two stages: The noise-identification stage and the filtering stage.

Literature on training DL models, with noisy or corrupted labels, suggests training a mentor/teacher network to filter samples or generate pseudo-labels for a student network to improve classification.[21] We partly employed the same strategy by training a reference model in the noise-identification stage to obtain reference risk scores. These reference risk scores were used, in the filtering stage, to exclude noise-inducing samples and create risk-stratified ensemble folds for the ensemble texture model. The noise-identification stage consisted of four steps:

1. Cancers were randomly matched with 20 healthy women on age (± one year age difference).
2. Cancers were then randomly split into five ensemble folds. In practice, we sampled the folds using a cross validation method stratified by age and cancer groups (SDC/IC/LTC/healthy). Note that the model performance was not cross validated per se as the folds were only used for training an ensemble model. The matched healthy women were added to the fold of their matched cancer.
3. Training views with a diagnosed cancer (SDC, IC, and LTCs) were excluded. For the matched healthy women, the same views were excluded, e.g., given a woman with cancer in the left breast, the two left views were excluded for both the woman with cancer and the matched healthy women. Unmatched healthy women were randomly split into the five validation sets.

No validation views were excluded. Consequently, the reference texture model was trained on two cancer contra-lateral views and validated on all four available views.

The reference risk scores were extracted after training the reference model for 40 epochs. The training procedure is described in section E below. This stage was applied to both the Dutch and Danish training datasets such that reference risk scores were available for all women in the two datasets.

The filtering stage, shown in Fig. 1c, was fundamentally like the noise-identification stage except for the sample filtering and ensemble fold sampling. Initially (before step 1), 10% of women healthy throughout follow-up with the highest reference risk were excluded. These were likely noisy samples as they might have a high PMD or texture risk and did not develop a breast cancer within follow-up. The SDCs, ICs, and LTCs with the lowest 4% reference risk, in each cancer group, were excluded as well because long-term risk estimation is inherently difficult hence a subset of women with a future cancer will be difficult to classify. Women with a future cancer could have low-risk tissue patterns that correlate highly with tissues of healthy women or have a cancer subtype that does not correlate with texture. In step 2, the ensemble folds were resampled and stratified on reference risk, cancer group, and age.

Unmatched/excluded women were split randomly, stratified on cancer group and reference risk, into the five validation sets. This filtering stage produced five folds for training a robust and better converging ensemble texture model (see Fig. 5).

*3.3 Augmentation-based Domain Adaptation*

A clinically relevant texture model must adapt to new target domains of unseen processing types. There are several methods for domain adaptation including data augmentation and adversarial trainings.[22,23] Adversarial approaches can be difficult to train in practice and often leads to failed convergence.[18,24] When the target domain is well defined, in our case by classical image transformations, adaptation can be successfully achieved by data augmentation which outperform adversarial approaches in certain cases.[22] To realize augmentation of a single raw view to multiple processed views of different formats, we developed a tool consisting of a series of four parameterized image transformations steps inspired by previous literature.[25,26]

In the first step, seen in Fig. 2a, a mask $M$ delineating the breast tissue was generated along with a distance map $D$ measuring distance to the skin-air-boundary. The background defined by $M$ was set to 0 which also removes burned-in view annotations.

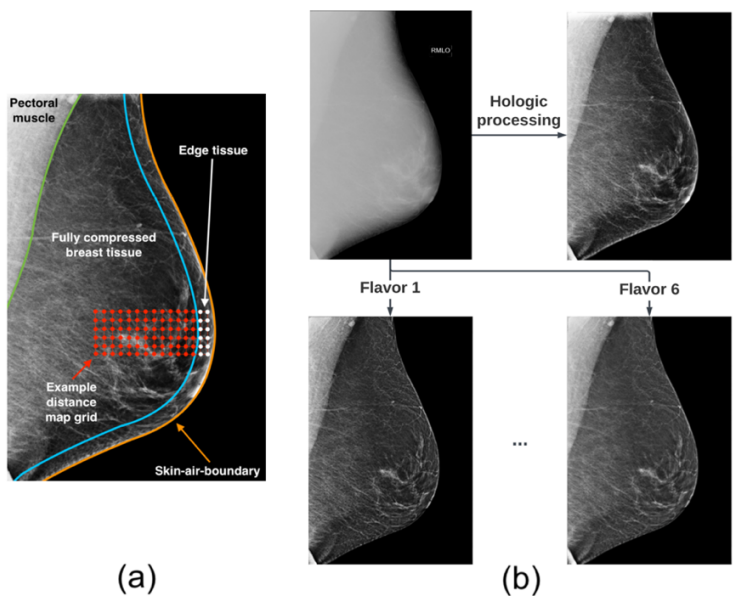

**Fig. 2. (a)** Example of the breast mask M and distance map D. Note that the rest of the grid has been omitted for viewing purposes and the grid and lines are not to scale. The tissue between the blue and orange lines is the edge tissue. **(b)** Example of a raw and a Hologic-processed view. Below is an example of a flavor 1 and 6 view. Flavor 2 to 5 views have been omitted in this example.

In the second step, pixel values in the breast tissue in view $I$ at position $i,j$ were transformed by a logarithmic and an inverse sigmoid function

$$I_{i,j} = \frac{I_{max}}{1 + \exp\left(\alpha\left(\log_{10}(I_{i,j}/\bar{I}_{breast})^2 - 1\right) + \beta\right)}, \quad (1)$$

where $\bar{I}_{breast}$ was the mean pixel value in the fully compressed breast tissue. The parameters $\alpha$ and $\beta$, were manually chosen parameters between values 4 to 5.5, and 0.7 to 1.2, respectively. The max value of a processed view $I_{max}$ was 4095.

In the third step, the edge tissue thickness was corrected for, and contrast was enhanced close to the skin-air-boundary as described by Tortajada et al..[25]

In the fourth step, the fully compressed breast tissue was contrast enhanced by addition of an enhancement mask as described by Panetta et al..[26]

$$I_{i,j} = \gamma \cdot I_{i,j} + \delta \cdot \frac{F_{i,j}}{I_{max}} \cdot I_{i,j}, \qquad (2)$$

where $F_{i,j}$ was the enhancement mask, which in practice was the difference between the original view and the low pass filtered view. Scaling parameters $\gamma$ and $\delta$, were manually tuned to approximate processed views from different vendors. A set of parameters is referred to as a flavor profile. The output of the tool, defined by the flavor profile, is a processed view which we refer to as a flavor. We manually constructed seven flavor profiles producing seven different flavors. The Danish and Dutch training datasets were both processed into six different flavors and the Danish testing dataset was processed into a seventh flavor. Consequently, the texture model was validated in a flavor not seen during training. Fig. 2b shows an example of a raw view, the corresponding Hologic-processed view, and examples of flavors created with the image conversion tool.

## 3.4 Training Procedure for the Texture Model

The texture model was generally trained using the same procedure and only the augmentation scheme and training/validation dataset varied. The texture model takes as input a single mammographic view, passes it through a ResNet18 encoder[27], initialized with pre-trained weights from the ImageNet challenge dataset[28], followed by two fully connected layers that output a single continues confidence measure. The model architecture is shown in Fig. 3.

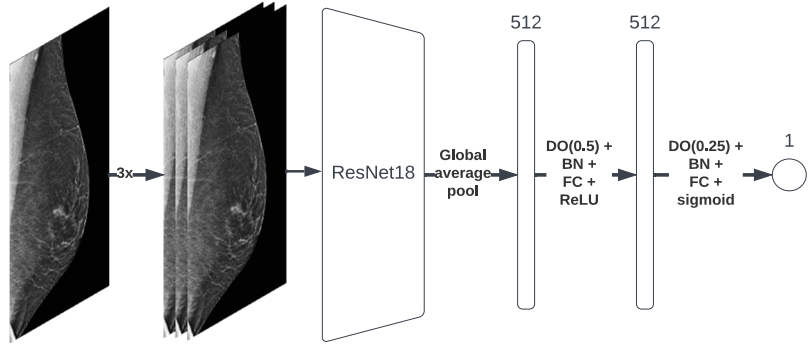

**Fig. 3.** The texture model architecture. The input view was repeated thrice in the channel axis and passed through ResNet18. The output feature map was global-average-pooled to a feature map of size 512, which was passed through a dropout (DO) layer with a probability of dropout of 0.5, a batch normalization (BN) layer, and a fully connected layer (FC) using, ReLU activation, to 512 hidden neurons. The last layers were a DO layer with a probability of 0.25, BN, and a FC layer with sigmoid activation to a single output neuron.

The model was trained to distinguish between cancer contra-lateral views of women with a future cancer and views from healthy women, by minimizing the binary cross entropy (BCE) in batch $B$ of size 10

$$BCE = -\frac{1}{|B|} \sum_{i=1}^{B} y_i \log(\hat{y}_i) + (1 - y_i) \log(1 - \hat{y}_i) ,$$
(3)

where $y$ is the binary ground-truth label and $\hat{y}$ is the scaled confidence generated by the model. The model was trained using Adam optimizer[29] with a learning rate of $10^{-5}$. During training right views are flipped across the vertical axis. Training views were randomly augmented to mimic difference in scanner positions by randomly rotating (by -15° to 15°), scaling (by a factor -0.2 to 0.2), or shearing in the horizontal direction (by a factor of -0.15 to 0.15).

The final texture model was an ensemble of five instances of the above-mentioned model trained using the same parameters. These five instances were trained on the corresponding five splits defined in the given curated training dataset. Each instance of the model was trained until convergence in the corresponding validation set indicated by the highest validation AUC where all cancer groups were considered as positives. All views, regardless of training or inference were scaled to a resolution of 0.255 mm per pixel and a size of padded to a size of 1194 by 938 pixels.

During inference on a testing dataset, the view-level texture risk score is an average of five confidence measures provided by each instance of model. Screened women had four views captured so to obtain a study-level risk score the four view-level risk scores were averaged. During inference, views in the given testing dataset were standardized to have the same mean and standard deviation as the used training dataset based on 1,000 randomly sampled views from the given testing dataset. In some experiments, the flavor augmentation was used. In these cases, all views in the given training dataset were processed into six flavors and randomly sampled into each batch. An epoch was defined as an iteration through 1/6 of the full training dataset and such that the model was trained on all flavor-augmented training samples by 6 epochs. The experiment section specifies in which experiment the flavor augmentation was used. All models have been trained and tested using Python (version 3.6.8) and TensorFlow (version 2.1.0) on Nvidia TITAN RTX GPUs.

## 4 Experiments

As described in the data curation section (Sec. 3.2), two reference texture models were initially trained to reduce noise and stabilize training. Meanwhile, this also provided an opportunity to analyze convergence before during and after the noise-identification and filtering stages. These convergence plots are shown in Fig. 5.

Afterwards, four series of experiments were conducted, referred to as A, B, C, and D, to establish a baseline performance using raw views, and assess whether the flavor augmentation ensured proper adaptation to domains of flavorized and Hologic-processed views. We estimated whether additional training data was needed and whether age, PMD, and presence of clips (a surrogate for a previous false-positive screen), could improve risk estimation. Fig. 4 depicts these fours series of experiments in a flow diagram. Results were denoted $R_{\text{train-format} \to \text{test-format}}^{\text{train-dataset} \to \text{test-dataset}}$, e.g., $R_{\text{flav} \to \text{proc}}^{\text{D-tr} \to \text{N-tst}}$ denoted the results of applying the texture model, trained on Danish training dataset with flavor views, to the NL processed testing dataset with Hologic-processed views.

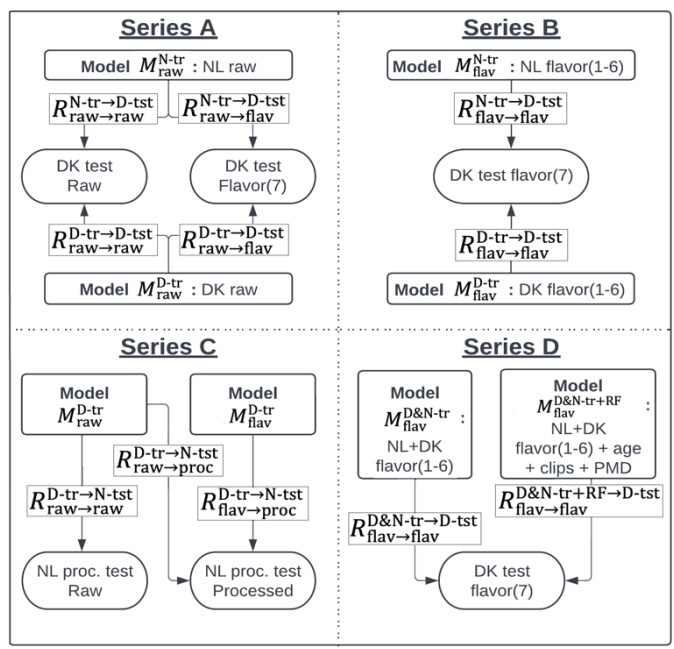

**Fig. 4.** Flow diagram of the four series of experiments. Boxes denote trained models, the used training dataset, and view formats. The oval endpoints denote the testing dataset used and view format. DK test=Danish testing dataset, NL proc. test=Dutch processed testing dataset, RF=risk factors.

The first series of experiments (see Fig. 4, series A) quantified a baseline performance in the Danish testing dataset of raw Siemens views using a texture model trained on raw Hologic or Siemens views. We also assessed whether the model trained on raw views, without flavor augmentation, would be able to generalize to flavor views.

In second series of experiments (see Fig. 4, series B), we quantified how well a texture model, now trained with flavor augmentation, would generalize to an unseen flavor (flavor 7) using the Dutch and the Danish training dataset of flavor views (flavor 1 to 6), respectively. Essentially, we simulated an unseen scanner.

In the third series of experiments (see Fig. 4, series C), baseline performance in the Dutch processed testing dataset was measured in raw views with the Danish trained model. We compared the baseline to the performance of the texture model trained with and without flavor augmentation to assess the performance on Hologic-processed views.

In the fourth series of experiments (see Fig. 4, series D), a texture model was trained on both the Dutch and Danish training dataset simultaneously to assess whether the training dataset size was sufficient. A significantly higher AUC would indicate the model was not saturated in terms of data. We concatenated the five ensemble sets from each of the training datasets and trained five new instances of the model using flavor views (flavor 1 to 6). This model was applied to the Danish testing dataset of flavor views (flavor 7). To assess whether risk factors could contribute to a full risk model, we separately trained a small neural network of three fully connected layers to combine the texture, age, presence of clips, and PMD.

Finally, we assessed whether the texture model was able to identify high-risk subsets of women in the Danish testing dataset. First, we reported how many women with ICs and LTCs were in the group of women with the 10% highest risks using $R_{\text{flav}\to\text{flav}}^{\text{D\&N-tr+RF}\to\text{D-tst}}$. Second, we reported sensitivities for ICs and LTCs at 90% specificity using $R_{\text{flav}\to\text{flav}}^{\text{D\&N-tr+RF}\to\text{D-tst}}$ as well. Third, the Danish testing dataset was split by texture risk (using $R_{\text{flav}\to\text{flav}}^{\text{D\&N-tr}\to\text{D-tst}}$) and PMD quantiles into 16 bins (4x4 quantiles) to assess the distribution and which of these bins might be flagged as high-risk. Quantiles were computed using healthy women only.

## 5  Results

In the following sections, the results are presented. Initially, the convergence of the reference texture model was investigated before and after the filtering stage. Next, the results of the four series of experiments are presented along with appropriate significance testing. Lastly, the percentages of ICs and LTCs in the high-risk groups were presented along with sensitivities at 90% specificity and the relationship between texture and PMD quantiles and odds ratios (OR) for each of the 16 bins.

*5.1 Performance of the Reference Texture Model*

Fig. 5 shows the convergence of the Danish reference model trained for 40 epochs during the noise-identification stage (top row) and after the filtering stage (bottom row).

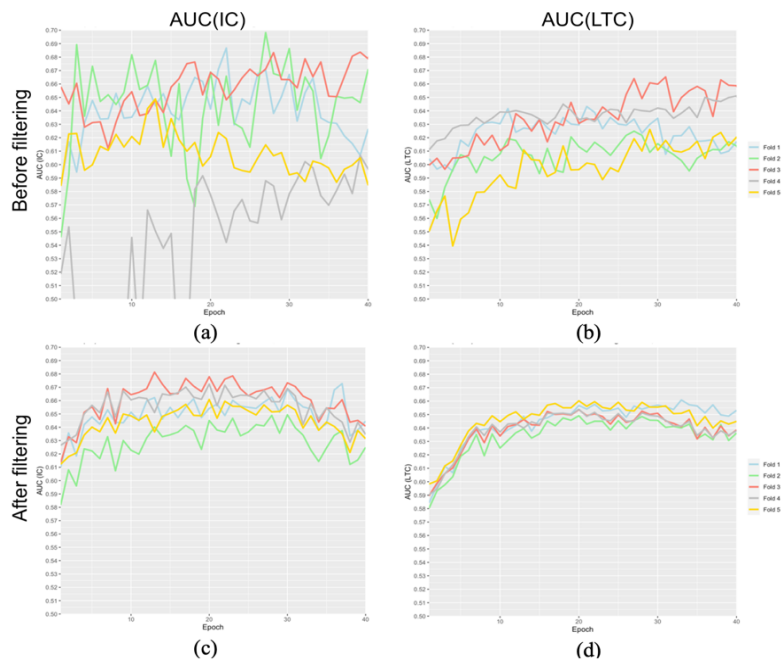

**Fig. 5.** Validation fold AUCs during training on raw views from the Danish training dataset. The top row is the AUCs for **(a)** ICs **(b)** and LTCs, during the noise-identification stage. The bottom row is the AUCs for ICs **(c)** and LTCs **(d)** after the filtering stage.

In the top row, the validation performances depended highly on when the training is stopped. Before the filtering stage, the folds converged, in terms of ICs, between epoch 14 and 39. The difference in AUC(IC) at epoch 28, the mean convergence point (CP), was 0.10. After the filtering stage, the largest difference in AUC(IC) at CP was 0.04. Before the filtering stage, the folds converged, in terms of LTCs, at epoch 21, 27, 31, 40, and 29, respectively. The difference in AUC(LTC) at CP was 0.05. After the filtering stage, fold one to five converged, in terms of LTCs, at epoch 33, 20, 20, 20, and 20, respectively. When evaluated at CP, the largest difference was 0.01 between folds. For both ICs and LTCs the validation curves were much closer together and converged largely at the same time after the two stages. Even with this curation of a training data, the folds overfit and does not guarantee equal convergence points, however, this effect was lessened by averaging over five instances of the ensemble texture model during inference.

*5.2 Future Breast Cancer Detection Performance*

Risk estimation performances, presented in Table 1, was measured using AUC.

Table 1: Texture model AUCs (95% CI) for ICs, LTCs, and ICs and LTCs combined for all experiments

| Experiment | AUC (IC) | AUC (LTC) | AUC (IC \| LTC) |
|---|---|---|---|
| **Series A** | | | |
| $R_{\text{raw}\to\text{raw}}^{\text{N-tr}\to\text{D-tst}}$ | 0.68 (0.65, 0.72) | 0.64 (0.62, 0.66) | 0.65 (0.63, 0.67) |
| $R_{\text{raw}\to\text{raw}}^{\text{D-tr}\to\text{D-tst}}$ | 0.70 (0.66, 0.74) | 0.65 (0.62, 0.67) | 0.66 (0.64, 0.68) |
| $R_{\text{raw}\to\text{flav}}^{\text{N-tr}\to\text{D-tst}}$ | 0.55 (0.50, 0.59) | 0.51 (0.49, 0.53) | 0.52 (0.50, 0.54) |
| $R_{\text{raw}\to\text{flav}}^{\text{D-tr}\to\text{D-tst}}$ | 0.52 (0.48, 0.57) | 0.52 (0.50, 0.55) | 0.51 (0.49, 0.53) |
| **Series B** | | | |
| $R_{\text{flav}\to\text{flav}}^{\text{N-tr}\to\text{D-tst}}$ | 0.69 (0.65, 0.73) | 0.64 (0.62, 0.66) | 0.65 (0.63, 0.67) |
| $R_{\text{flav}\to\text{flav}}^{\text{D-tr}\to\text{D-tst}}$ | 0.70 (0.66, 0.74) | 0.64 (0.62, 0.66) | 0.65 (0.64, 0.67) |
| **Series C** | | | |
| $R_{\text{raw}\to\text{raw}}^{\text{D-tr}\to\text{N-tst}}$ | 0.66 (0.61, 0.72) | 0.60 (0.58, 0.62) | 0.61 (0.59, 0.63) |
| $R_{\text{raw}\to\text{proc}}^{\text{D-tr}\to\text{N-tst}}$ | 0.53 (0.47, 0.59) | 0.55 (0.53, 0.57) | 0.54 (0.52, 0.56) |
| $R_{\text{flav}\to\text{proc}}^{\text{D-tr}\to\text{N-tst}}$ | 0.66 (0.60, 0.71) | 0.63 (0.61, 0.65) | 0.63 (0.61, 0.65) |
| **Series D** | | | |
| $R_{\text{flav}\to\text{flav}}^{\text{D\&N-tr}\to\text{D-tst}}$ | 0.71 (0.67, 0.75) | 0.65 (0.63, 0.67) | 0.66 (0.64, 0.68) |
| $R_{\text{flav}\to\text{flav}}^{\text{D\&N-tr+RF}\to\text{D-tst}}$ | 0.71 (0.67, 0.75) | 0.68 (0.65, 0.70) | 0.68 (0.66, 0.70) |

For AUC(LTC) only LTCs were considered positive. For AUC(IC) only ICs were considered positive. For completeness, we included AUC(IC | LTC) where both ICs and LTCs were considered positives. All women not considered positive were negative. A paired (unless stated

otherwise) two-sided DeLong test was used to determine whether two AUCs were significantly different and used to compute 95% confidence intervals (CI).[30]

In series A, the baseline results $R_{\text{raw}\to\text{raw}}^{\text{N-tr}\to\text{D-tst}}$ and $R_{\text{raw}\to\text{raw}}^{\text{D-tr}\to\text{D-tst}}$ yielded AUCs(IC) of 0.68 and 0.70, respectively. The AUCs(LTC) were 0.65 and 0.64, respectively. There were no significant differences between the AUCs(IC) or AUCs(LTC) with p(IC)=0.24 and p(LTC)=0.35, which indicated good adaptation across raw views and robust model training. $R_{\text{raw}\to\text{flav}}^{\text{N-tr}\to\text{D-tst}}$ and $R_{\text{raw}\to\text{flav}}^{\text{D-tr}\to\text{D-tst}}$ yielded near random performance that was significantly worse compared to baseline p<0.001 for ICs and LTCs.

In series B, the results $R_{\text{flav}\to\text{flav}}^{\text{N-tr}\to\text{D-tst}}$ and $R_{\text{flav}\to\text{flav}}^{\text{D-tr}\to\text{D-tst}}$ yielded AUCs(IC) of 0.69 and 0.70, respectively. The AUCs(LTCs) were both 0.64. There was no difference in AUC between these two results with p(IC)=0.24 and p(LTC)=0.9. $R_{\text{flav}\to\text{flav}}^{\text{N-tr}\to\text{D-tst}}$ and $R_{\text{flav}\to\text{flav}}^{\text{D-tr}\to\text{D-tst}}$ were not different from the corresponding baselines $R_{\text{raw}\to\text{raw}}^{\text{N-tr}\to\text{D-tst}}$ and $R_{\text{raw}\to\text{raw}}^{\text{N-tr}\to\text{D-tst}}$ with p(IC)=0.57/p(LTC)=0.64 and p(IC)=0.76/p(LTC)=0.39, respectively. This suggested that the texture models, trained with flavor augmentation, generalized to an unseen flavor domain without any loss of risk assessment performance.

In series C, the baseline results $R_{\text{raw}\to\text{raw}}^{\text{D-tr}\to\text{N-tst}}$ yielded an AUC(IC) of 0.66 and AUC(LTC) of 0.60, which is low due to difference in curation of datasets. $R_{\text{flav}\to\text{proc}}^{\text{D-tr}\to\text{N-tst}}$ yielded an AUC(IC) of 0.66 as well, but a significantly higher AUC(LTC) of 0.63 with p(LTC)<0.001, which indicated that the flavor model adapted better to unseen processed views than the baseline on raw views. $R_{\text{raw}\to\text{proc}}^{\text{D-tr}\to\text{NL-tst}}$ yielded near random performance which was significantly worse than the baseline with p<0.001 for ICs and LTCs. $R_{\text{flav}\to\text{proc}}^{\text{D-tr}\to\text{N-tst}}$ was not different from $R_{\text{flav}\to\text{flav}}^{\text{D-tr}\to\text{D-tst}}$ with p(IC)=0.17 and p(LTC)=0.42, using an unpaired test. This suggested that the texture model, trained with flavor augmentation, performed equally well when validated on flavorized Siemens views of Hologic-processed views.

In series D, $R_{\text{flav}\to\text{flav}}^{\text{D\&N-tr}\to\text{D-tst}}$ did not yield significantly higher AUCs than $R_{\text{flav}\to\text{flav}}^{\text{N-tr}\to\text{D-tst}}$, with p(IC)=0.06 and p(LTC)=0.14. However, $R_{\text{flav}\to\text{flav}}^{\text{D\&N-tr}\to\text{D-tst}}$ gave a higher AUC(LTC) than $R_{\text{flav}\to\text{flav}}^{\text{D-tr}\to\text{D-tst}}$ with p(LTC)=0.02. There was only partial evidence that doubling the training dataset size increased risk estimation performance. $R_{\text{flav}\to\text{flav}}^{\text{D\&N-tr+RF}\to\text{D-tst}}$ gave a significantly higher AUC(LTC) of 0.68 compared to $R_{\text{flav}\to\text{flav}}^{\text{D\&N-tr}\to\text{D-tst}}$, with p(LTC)<0.001 suggesting that established risk-factors, combined with texture in a small neural network, contributed with additional information on long-term risk.

*5.3 High-risk Subgroups and Distribution of Cancers Texture and PMD quantiles*

In the group of women in the Danish testing dataset with the highest risk from $R_{\text{flav}\to\text{flav}}^{\text{D\&N-tr+RF}\to\text{D-tst}}$, 25.5% of women had an IC and 24.8% had an LTC.

Using risk from $R_{\text{flav}\to\text{flav}}^{\text{D\&N-tr+RF}\to\text{D-tst}}$, the sensitivities at 90% specificity was 25.5% for ICs and 25.5% for LTCs.

Using texture risk from $R_{\text{flav}\to\text{flav}}^{\text{D\&N-tr}\to\text{D-tst}}$ and PMD quantiles of healthy women, three matrices were created using and shown in Fig. 6.

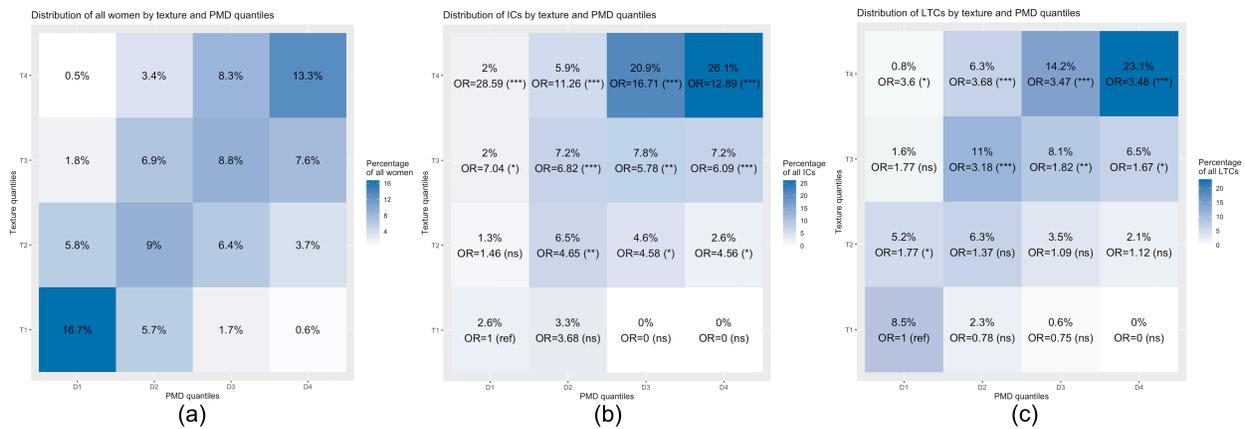

**Fig. 6.** Matrix plots of **(a)** the distribution of all women in the Danish testing dataset, **(b)** distribution of ICs, and **(c)** distribution of LTCs in texture and PMD quantiles. Texture scores were from $R_{\text{flav}\to\text{flav}}^{\text{D\&N-tr}\to\text{D-tst}}$. Quantiles were calculated in healthy women only. Underneath each percentage in (a) and (b), the odds ratios (OR) are specified when compared to the reference group and exact Fisher p-values are in parentheses.

The leftmost matrix (a) shows the distribution of all women, the middle matrix (b) shows the distribution of ICs, and the rightmost matrix (c) show the distribution of LTCs. Texture and PMD were correlated as most cancers were found near the diagonal and in the top right corner in all matrices. The highest ORs found in both matrices were in the top rows. This indicated that, at any PMD quantile, being in the fourth texture quantile yielded a significantly higher OR of having an IC or LTC compared to the reference group. Women in (D3, T4) corresponded to 8.3% of women and accounted for 20.9% of ICs and 14.2% of LTCs. However, simply selecting the top 8.3% of women with highest risks, using $R_{\text{flav}\to\text{flav}}^{\text{D\&N-tr+RF}\to\text{D-tst}}$, accounted for 24.2% and 22.1% of ICs and LTCs, respectively.

Notably, the highest ORs for each row are not in D4 column, which indicate that high-density women are not at the highest risk when simultaneously considering texture. This effect seems most pronounced for LTCs where the highest ORs are located in the D1 or D2 column.

## 6 Discussion

In this study, we developed a clinically usable, cross-vendor, and robust texture model for long-term breast cancer risk. By employing a flavor augmentation-based domain adaptation technique, we enabled the texture model to adapt to the domain of unseen Hologic-processed views in a large independent dataset. The texture model was trained to rely on features of healthy breast tissue to optimally learn long-term risk features. The texture model was trained as an ensemble model on a curated training dataset that prevented erratic convergence and yielded a robust risk estimate.

*6.1 The Texture Model Domain Adaptation Abilities*

In the series A experiments, results suggested that the texture model adapted well across domains of raw views from different devices, and that the ensemble training gave consistent results when populations and devices changed. $R_{\text{raw}\to\text{flav}}^{\text{N-tr}\to\text{D-tst}}$, $R_{\text{raw}\to\text{flav}}^{\text{D-tr}\to\text{D-tst}}$, $R_{\text{raw}\to\text{proc}}^{\text{D-tr}\to\text{N-tst}}$ failed to adapt to processed views which indicated that a domain adaptation technique was needed to support a clinical workflow with processed views.

In the series B experiments, we simulated a new scanner by using a seventh and unseen flavor during testing. In this case the flavor texture models adapted equally well to an unseen flavor and performed equal to their raw baseline counterparts. This indicated that the augmentation-based domain adaptation technique succeeded in enabling the model to learn risk features in vendor-agnostic flavor views and might enable integration with the typical clinical workflow.

The series C experiments showed that the flavor augmentation enabled to train a texture model that adapted better to Hologic-processed views than its corresponding raw baseline. It also confirmed possible integration in a clinical workflow. $R_{\text{raw}\to\text{raw}}^{\text{D-tr}\to\text{N-tst}}$ suggested that there was a bias in which women from the Dutch population, for which both a raw and a Hologic-processed view were available. Series B and C experiment highlighted the gain of training on flavorized raw views. There was no need to collect large datasets of processed views across several sites and scanners, as flavor augmentation successfully enabled training a cross-vendor texture model using augmented single site data only.

The series D experiment showed only little evidence that doubling the dataset increased performance. These results indicated that the performance is close to the upper limit for this particular model. However, adding age, clips, and PMD yielded better long-term risk assessment. If used in a clinical workflow this best performing full risk model should be used.

*6.2 Comparison to Existing Risk Models*

We compared the texture model to traditional models for estimating risk. The Gail model achieved an AUC of 0.60 (95% CI 0.58 to 0.62) across 29 datasets in 5 to 10-year risk estimation.[31] The TC model achieved, in 35,921 women with six years follow-up, an AUC of 0.62 (95% CI 0.60 to 0.64).[32]

The BOADICEA model included all established risk factors: personal, familial, reproductive, hormonal, and lifestyle factors along with polygenic risk scores, breast density, and pathogenic variants. It achieved an AUC of 0.70 (95% CI 0.66 to 0.73) for five-year risk in 5,693 women.[33] Our texture model achieved a considerably higher AUC than the Gail and TC model, but the BOADICEA model achieved a slightly higher five-year AUC than our model that yielded 0.68 (95% CI 0.66 to 0.70), however likely not significantly so, and the study sample was less than a tenth the size of ours. The texture model, however, has the advantage using screen-available information only as opposed to collecting patient history, gene data, and questionnaires.

A. Gastounioti *et al.* used iCAD to achieve an AUC of 0.68 (95% CI 0.64-0.72) in detecting cancers for 5,139 screen negative women with up to two years after screening.[13] Comparably, we achieved an AUC(IC) of 0.71 (95% CI 0.67 to 0.75) for the same interval after screening (IC).

Kallenberg et. al achieved an AUC of 0.61 (95% CI 0.57 to 0.66) using a texture model.[15] Our texture model was inspired by this study and results indicate a clear improvement over the previous results.

Dembrower *et al.* achieved an AUC of 0.65 (95% CI 0.63 to 0.66) one- to six-year risk estimation in a dataset of 2,283 women.[17] This AUC corresponded well with our results: 0.65 (95% CI 0.63 to 0.67). However, cancers diagnosed close to screening are generally easier to detect, as indicated by our texture model's AUC(IC) of 0.71. Consequently, the AUC for our texture model could be hypothesized to increase if we evaluated risk after one year after screening as well.

Mirai obtained AUCs between 0.68 to 0.73 in 6 months to five-year risk estimation in seven validation datasets.[18] These results are not comparable to ours, as the authors considers all cancers six months to five years after screening as positives and 42% to 82% of cancers were diagnosed in the first year. This distribution of time from screening to diagnosis might not be representative of a population-based screening program.[20]

Neither of the four studies on DL risk models assessed adaptation to unseen vendor-processed views.

*6.3 Clinical Relevance*

For both ICs and LTCs it was observed, in Fig. 6, that being in the fourth texture quantile, at any PMD quantiles, yielded the largest ORs compared to the reference group. This result showed that texture consistently contributed to better risk assessment besides PMD. Women in third PMD quantile and fourth texture quantile could be flagged as high-risk.

It turned out that when selecting 8.3% of the women with the highest risk, using the texture risk combined with established risk factors, 24.2% of ICs and 22.1% of LTCs were detected. These percentages were higher than the percentages when choosing women in (D3, T4) which also corresponded to 8.3% of the women, but only accounted for 20.9% of ICs and 14.2% of LTCs. This suggest that, in a clinical workflow, simply choosing women with the highest risk is more effective that using texture/PMD quantiles (Fig. 6) to flag women as high-risk. Using texture/PMD quantiles, we observed that the highest ORs were not found in the D4 column indicating that supplementary screening strictly for high-density women[5] may not be optimal if texture was considered as well. A large proportion of persistently healthy women, with low texture risk, will unnecessarily receive supplemental screening while some high-risk high-texture women will be missed. However, it is well-known that screening sensitivity is lower for women with high breast density due to a masking effect.[5] It may therefore still be valuable for clinicians to consider PMD along with texture risk.

*6.4 Limitations*

Although we successfully developed a texture model, our study was limited by using only one testing dataset of vendor-processed view, which was inherently different from the Danish consecutive screening cohort due to differences in curation. We cannot accurately assess whether six flavors for training was sufficient to generalize to other vendors though results suggest it. The study was limited by the similarity of the populations, as they had the same screening intervals, number of readers, and were demographically comparable.

*6.5 Conclusion and Future Perspectives*

We successfully identified a clinically relevant high-risk cohort using a mammographic texture model for long-term risk. We would like to investigate how to incorporate it into screening workflow in collaboration with radiologists, and whether texture correlates with certain subtypes of breast cancer. The robust cross-vendor texture model should further be validated in a cohort with longer follow-up, more demographic diversity, and with different screening intervals to confirm measured performance and robustness. Lastly, we want to supplement the texture model, that estimates long-term risk, with a detection model, that estimate immediate/short-term risk using localized findings or precursors. Combining two such risk models, after being trained optimally and separately, might increase short- and long-term risk estimation, as no weak risk signals are suppressed.

**Disclosures**

None of the authors have anything to disclose.

**Acknowledgements**

This work was partly supported by Eurostars (grant E9714 IBSCREEN) and was approved by the Danish Data Inspection Agency and Danish Patient Safety Authority (reference number 3–3013–2118).